\def\year{2022}\relax
\documentclass[letterpaper]{article} 
\usepackage{aaai22}  
\usepackage{times}  
\usepackage{helvet}  
\usepackage{courier}  
\usepackage[hyphens]{url}  
\usepackage{graphicx} 
\usepackage{natbib}  
\usepackage{caption} 

\usepackage{times}
\usepackage{epsfig}
\usepackage{graphicx}
\usepackage{amsmath}
\usepackage{amssymb}
\usepackage{booktabs}
 \usepackage{multirow}
\usepackage{float}
\usepackage{bbding}
\usepackage{indentfirst}
\usepackage{cite}
\usepackage{multirow}

\DeclareCaptionStyle{ruled}{labelfont=normalfont,labelsep=colon,strut=off} 
\usepackage{algorithm}
\usepackage{algorithmic}

\usepackage{newfloat}
\usepackage{listings}
\floatstyle{ruled}
\newfloat{listing}{tb}{lst}{}
\floatname{listing}{Listing}

\title{Multi-scale frequency separation network for image deblurring}
\author {
    Yanni Zhang\textsuperscript{\rm 1,\thanks{Equal contribution authors. $^\dagger$Corresponding authors. }},
    Qiang Li \textsuperscript{\rm 1,$^*$},
    Miao Qi \textsuperscript{\rm 1},
    Di Liu \textsuperscript{\rm 1},
    Jun Kong\textsuperscript{\rm 1,2,$^\dagger$} and
    Jianzhong Wang \textsuperscript{\rm 1,$^\dagger$}
}
\affiliations {
   \textsuperscript{\rm 1} College of Information Science and Technology, Northeast Normal University, China\\
    \textsuperscript{\rm 2} Key Laboratory of Applied Statistics of MOE, Northeast Normal University, China
}

\usepackage{bibentry}

\begin{document}

\maketitle

\begin{abstract}
Image deblurring aims to restore the detailed texture information or structures from blurry images, which has become an indispensable step in many computer vision tasks. Although various methods have been proposed to deal with the image deblurring problem, most of them treated the blurry image as a whole and neglected the characteristics of different image frequencies. In this paper, we present a new method called multi-scale frequency separation network (MSFS-Net) for image deblurring. MSFS-Net introduces the frequency separation module (FSM) into an encoder-decoder network architecture to capture the low- and high-frequency information of image at multiple scales. Then, a cycle-consistency strategy and a contrastive learning module (CLM) are respectively designed to retain the low-frequency information and recover the high-frequency information during deblurring. At last, the features of different scales are fused by a cross-scale feature fusion module (CSFFM). Extensive experiments on benchmark datasets show that the proposed network achieves state-of-the-art performance.
\end{abstract}

\section{Introduction}
The blur artifact will affect the image quality and severely degrade the performance of downstream computer vision tasks, such as video surveillance, object detection and face recognition. Therefore, accurate and efficient image deburring techniques have attracted much attention in both academic and industrial communities.

In the early studies, most image deblurring methods focused on estimating the blurry kernel by introducing some prior information~\cite{41}. However, since the blur in an image may be induced by multiple reasons, image deblurring becomes a highly ill-posed problem and it is difficult to model the complex blur kernel by simple and linear assumptions.

\begin{figure}[t]
\centering
\includegraphics[width=0.9\columnwidth]{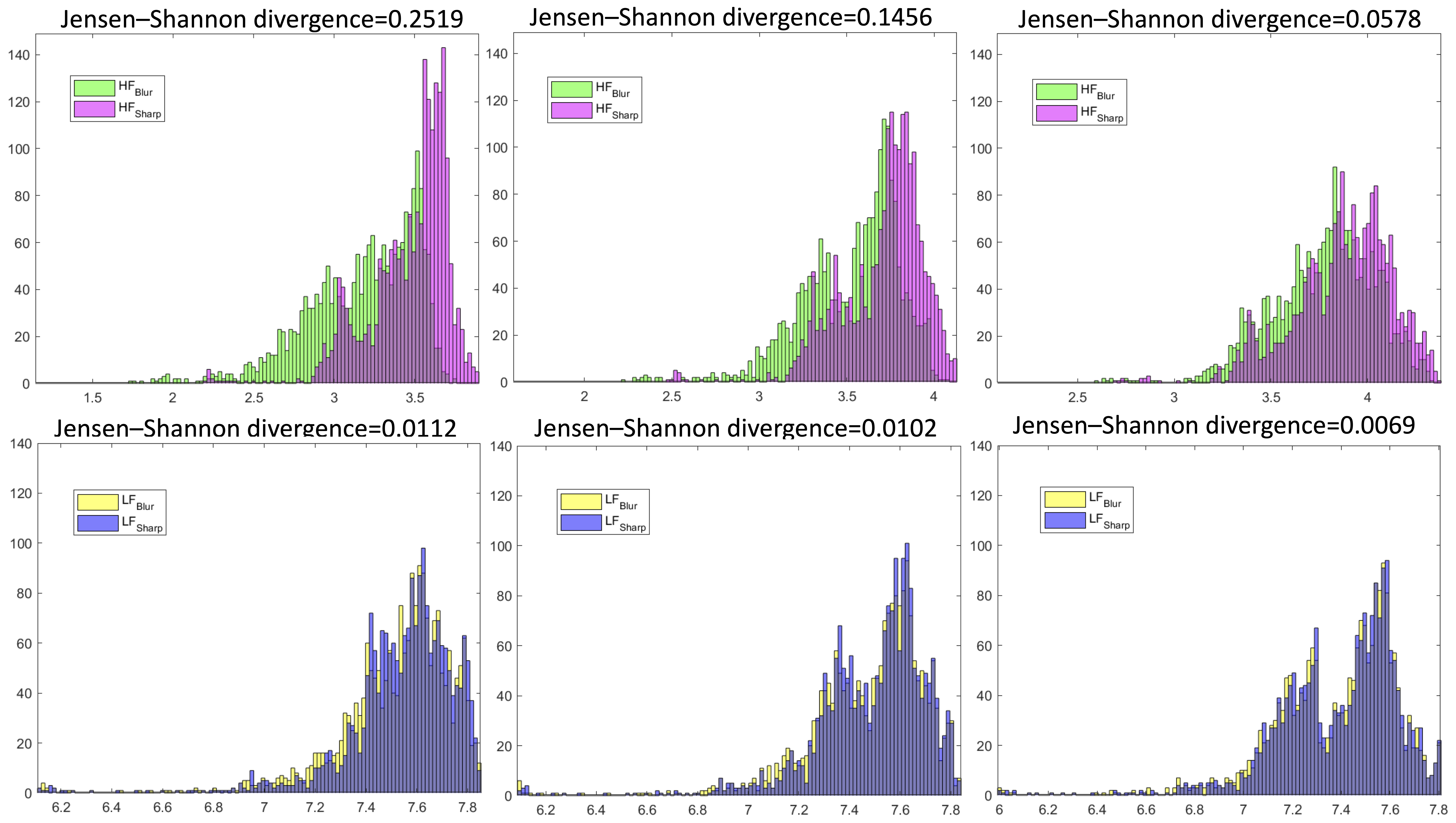} 
\caption{The distributions of entropy obtained by samples in GoPro dataset. In this figure, the LF component of each image is obtained by a low-pass Gaussian filter and the HF component is obtained by subtracting the LF component from the original image. Top: From left to right are the distributions of entropy obtained by HF components of sharp ($HF_{Sharp}$) and blurry ($HF_{Blur}$) images at original, 1/2 and 1/4 scales. Down: From left to right are the distributions of entropy obtained by LF components of sharp ($LF_{Sharp}$) and blurry ($LF_{Blur}$) images at original, 1/2 and 1/4 scales.}
\label{fig1}
\end{figure}

With the development of deep learning, some deep convolutional neural networks (CNNs) have been adopted as blur kernel estimator and showed satisfied deblurring performance~\cite{4}. However, these methods always need two stages (i.e. blur kernel estimation and blurry image decovolution) to accomplish the image deblurring task. Therefore, they may suffer from both high computational burden and inaccurate blur kernel estimation. More recently, some other CNN based image deblurring methods were proposed to directly learn the relationship between blurry and sharp images by an image-to-image regression manner~\cite{7,8,9,10,13}. Compared with other works, the advantage of image-to-image regression methods is that they could avoid the deblurring errors induced by inadequate blur kernel estimation. Besides, the CNN has also been combined with some other techniques such as recurrent neural network (RNN) and generative adversarial network (GAN) for image deblurring. 

Although the aforementioned CNN based deblurring methods adopted various techniques to remove the blur from images, most of them leveraged the encoder-decoder architecture to capture multi-scale image features. That is, they first leverage encoder to gradually reduce the input blurry image to low-resolution representations, and then utilize decoder to progressively recover the original resolution for deblurring. This multi-scale strategy is reasonable for image deblurring because the low-resolution representations can easily capture coarse image features while the high-resolution representations are more suitable to recover fine image details. However, the differences of image information not only exist in the resolution scales but also can be reflected by different frequencies. That is, the smoothly changing structure and outline of an image are mainly described by its low-frequency (LF) component, while the fine details with rapid variations in image are usually described by its high-frequency (HF) component. Therefore, since the existing CNN based image deblurring methods dealt the image feature at each scale as a whole and neglected to distinguish image frequencies, their performance may not be optimal. 

To overcome the above limitation, we propose a multi-scale frequency separation network (MSFS-Net) for image deblurring. MSFS-Net combines the multi-scale strategy with a frequency separation module (FSM) to capture different image features from both resolution scale and frequency aspects. Furthermore, different frequency information of the image is processed discriminatively in our work. Specifically, a simple cycle-consistency criteion is employed to maintain the LF features and a contrastive learning based module is proposed to progressively restore the HF features at different scales. Finally, a cross-scale feature fusion module (CSFFM) is also designed to compensate the information loss caused by the down-sampling of resolution scale and better fuse the feature of different scales. Experimental results and ablation analysis on three benchmark datasets demonstrate that with the help of frequency separation module and other components in our method, the proposed MSFS-Net can achieve state-of-the-art performance.

Our main contributions are fourfold:

\begin{itemize}
\item We propose a frequency separation module (FSM) to divide the image features into LF and HF components. Through embedding FSM into an encoder-decoder network architecture, our MSFS-Net can comprehensively capture image features of different frequencies and scales.
\item To differentially deal with the various features, a cycle-consistency strategy and a contrastive learning module (CLM) are proposed to constrain the LF and HF features, respectively.
\item We propose a cross-scale feature fusion module (CSFFM) to fuse the features of encoder and decoder from different scales, so that the multi-scale information can be better used to facilitate the deblurring.
\item Extensive experiments are conducted to demonstrate the effectiveness of our proposed MSFS-Net and the modules in it.
\end{itemize}

\section{Related Works}

\subsection{Image Deblurring}

Nowadays, the deep CNN models with image-to-image regression strategy have been proved to be effective for image deblurring task. The pioneer work was multi-scale CNN (MSCNN) proposed by~\cite{7}. Inspired by MSCNN, ~\cite{23} proposed a coarse-to-fine image deblurring network with selective parameter sharing and nested skip connections between different sub-networks.~\cite{24} adopted an encoder-decoder backbone with dense deformable and self-attention modules to improve the deblurring performance.~\cite{8} presented a multi-input multi-output U-net (MIMO-UNet) which utilized a single U-Net (i.e., encoder-decoder with short connections) but multiple input and output images to handle the coarse-to-fine image deblurring.~\cite{25} utilized an encoder-decoder network to extract multi-scale image features, and then integrated the auxiliary and meta learning to enhance the deblurring performance.~\cite{15} also applied encoder-decoder architecture to implement multi-scale and multi-stage image restoration tasks by introducing a new normalization method. In order to achieve better deblurring effect,~\cite{17} proposed an image deblurring approach by utilizing RNN to receive different directional sequence of CNN features.~\cite{10}  proposed a scale-recurrent network (SRN) by introducing the long-short term memory (LSTM) and ResBlock into an encoder-decoder based deblurring model. The success of GAN also promoted image deblurring research.~\cite{19} proposed a DeblurGAN to model different blur sources, in which a CNN with encoder-decoder architecture was employed as generator and a convolutional patch-based classifier was adopted as discriminator. Based on DeblurGAN, DeblurGAN-v2~\cite{20} was proposed to incorporate a double-scale discriminator and a feature pyramid network into GAN to achieve better deblurring result. 

\subsection{Frequency Separation}

An image can be decomposed into different frequency bands, and different frequency bands contain structures and textures with distinct characteristics. Therefore, analyzing the image feature in frequency domain is a commonly used technique in many conventional low-level computer vision tasks. Recently, researchers have also proposed some deep learning based deblurring methods which consider the characteristics of different image frequency.~\cite{2,3} employed CNN for blur kernel estimation in frequency domain. In image-to-image regression framework,~\cite{13} designed a two-stage method which first separates the HF residual information from the blurry image and then adopts an encoder-decoder network to realize the information refinement.~\cite{9} utilized discrete wavelet transform to divide the dilated convolution features into four frequency bands, so that different frequency features can be refined independently. Nevertheless, the above two methods only separated the image frequency in the first or last layer of the network. Thus, they can only capture the image features of different frequencies from a specific scale and ignored the different image frequency features of multiple scales. 

\subsection{Contrastive Learning}

Contrastive learning~\cite{29} is a widely used self-supervised strategy. Motivated by the success of its application in representation learning~\cite{31}, some researchers have adopted contrastive learning to model the comparative relationships between features in computer vision tasks~\cite{34,35}. Recently,~\cite{38} adopted contrastive learning in an image-to-image translation network.~\cite{39} designed a network using contrastive learning to remove the haze from hazy image.~\cite{40} also applied contrastive learning to obtain invariant degradation representation in image super-resolution problem. Although these methods demonstrated that contrastive learning can help to improve the performance of some low-level vision tasks, there are few works employ contrastive learning in image deblurring problem. Therefore, how to make good use of contrastive learning to facilitate the performance of image deblurring is still needed to be studied.

\section{Method}

\subsection{Motivation}

In multi-scale and hierarchical image deblurring methods, researchers have realized that the image features of different scales or spatial resolutions reflect diverse characteristics~\cite{7}. Nevertheless, previous deblurring works seldom took the frequency information of image into consideration. In this study, we observe that the difference between blurry and sharp images lies in both the scale and frequency aspects. To justify our observation, we compare the entropy obtained by all samples with different frequencies and scales in GoPro dataset. From the distributions in Fig. 1, we can find that the discrepancy between blurry and sharp images at the same scale is mainly reflected by their HF components. Specifically, the Jensen–Shannon divergences between entropy distributions obtained by HF components of sharp and blurry images are much larger than those obtained by LF components. This phenomenon may due to that blurring can be regarded as a process of diffusing the information encoded in sharp edges across an image, which would not dramatically alter the smoothly changing structure and outline of the image~\cite{41}. Moreover, we can also see that the difference between entropy distributions of sharp and blurry images with large scale is greater than that with small image scale since the down-sampling will sacrifice the texture details of images.

\subsection{Overview}
Motivated by the observation from Fig. 1, we propose a multi-scale frequency separation network (MSFS-Net), which makes full use of different frequency features at different scales, to achieve better deblurring performance. Figure 2 shows the overall architecture of the MSFS-Net.

\begin{figure}[t]
\centering
\includegraphics[width=0.9\columnwidth]{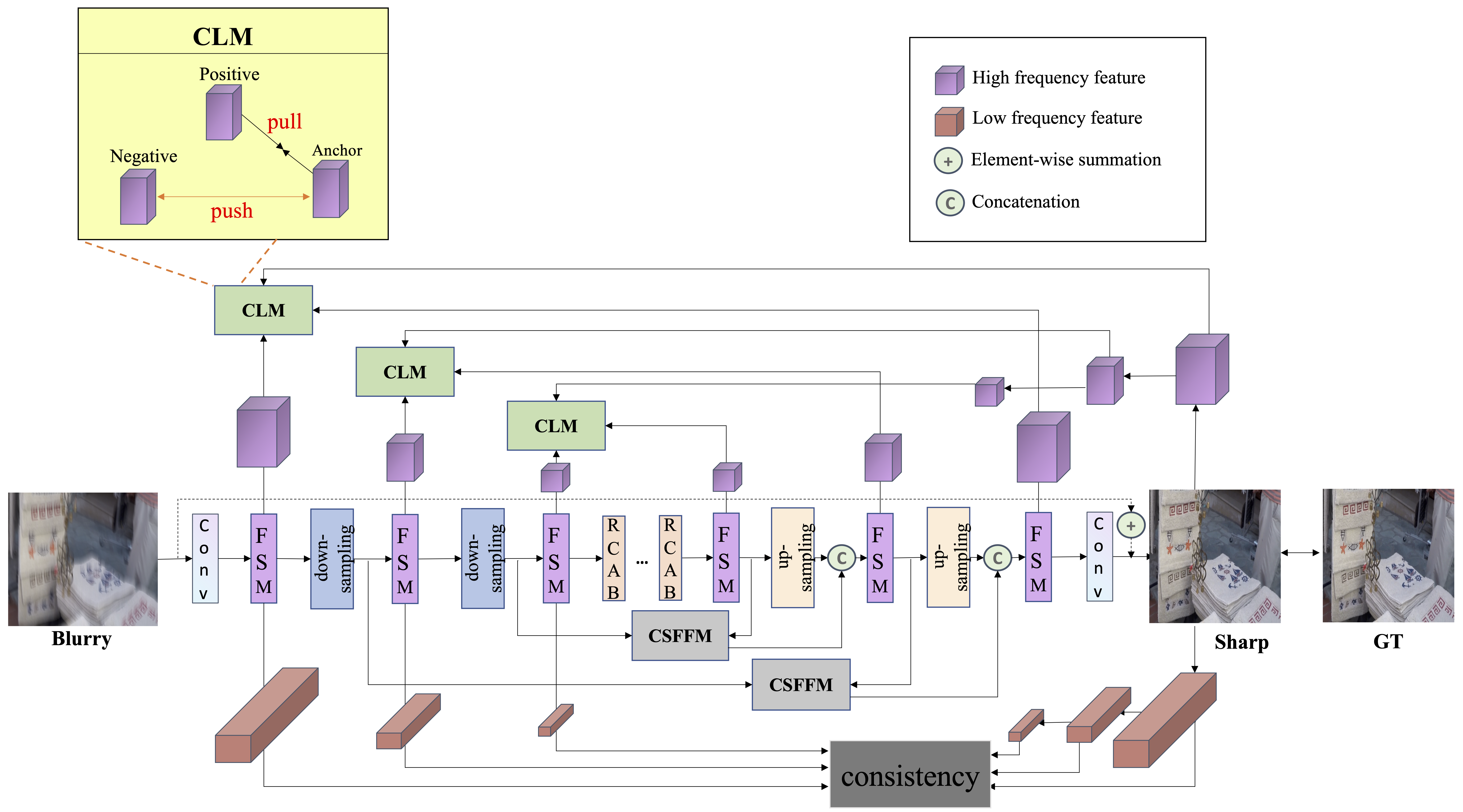} 
\caption{The architecture of the proposed MSFS-Net.}
\label{fig1}
\end{figure}

As can be seen from Fig. 2, the architecture of MSFS-Net is based on an encoder-decoder structure to hierarchically extract multi-scale image features. First, a blurry image is input and a 3$\times$3 convolution is applied to get shallow features. Then, the down-sampling module and frequency separation module (FSM) are combined in the encoder stage to progressively extract the LF and HF features of image at different scales. The down-sampling module consists of 3$\times$3 convolution with step 2 and LeakyRelu, and FSM is proposed to decompose the down-sampled features into different frequency. After the encoder stage, multiple RCABs~\cite{43} are further adopted to refine the latent feature and improve the model capacity. Next, we use up-sampling module to achieve scale restoration of features in the decoder stage. The up-sampling module consists of RCAB with pixel-shuffle~\cite{43} and FSM is also adopted to decompose the restored features at each scale. Since the decoder stage requires delicately use of fine-grained details to reconstruct features, the cross-scale feature fusion module (CSFFM) is applied to connect features at different scales of encoder and decoder stages so that different context information can be passed to each other and well preserved. In order to minimize the loss of information and make the network converge rapidly, we fuse the feature of input image with the features after the last 3$\times$3 convolution of decoder by an element-wise summation. Last but most important, in order to take full advantage of LF and HF information, we reuse the encoder of network to obtain different frequency features of output sharp image at different scales, and two distinct strategies are carried out to constrain the LF and HF features in the intermedia layers of our network. On the one hand, since LF features of the blurry and sharp images at the same scale are similar, a simple cycle-consistency criterion is utilized to ensure that the LF features of the output sharp and input blurry images are not far away from each other. On the other hand, we propose a contrastive learning module (CLM) to regularize the HF features in the decoder stage, so that the interference of HF features in blurry image can be effectively removed. Here, it should be noted that we utilize the LF and HF components of output sharp image to constrain the intermedia features of different stages in the backbone network (i.e., LF for encoder constraint and HF for decoder constraint). This is because that the encoder is mainly used to extract context and outline information of the blurry image while the detailed information of sharp image is mostly generated by the decoder. Moreover, the cycle-consistency and CLM introduce multiple closed-loop structure in our network, which is helpful to reduce the solution space of our model~\cite{44}.

\subsection{Frequency Separation Module}

\begin{figure}[h]
\centering
\includegraphics[width=0.9\columnwidth]{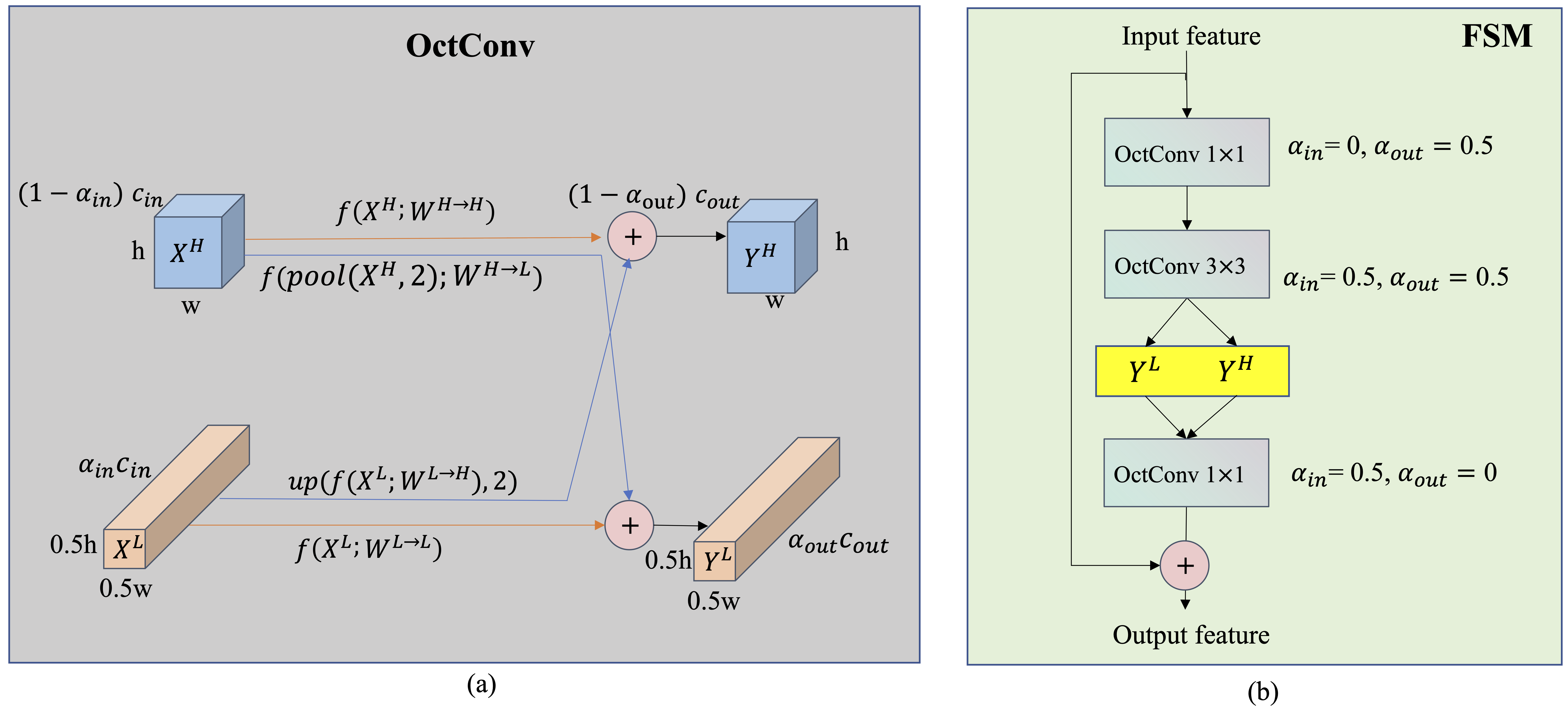} 
\caption{(a) The architecture of OctConv. (b) The architecture of frequency separation module (FSM).}
\label{fig1}
\end{figure}

Inspired by~\cite{45}, Octave Convolution (OctConv) is used as the basic block of our frequency separation module (FSM). The structure of OctConv is shown in Fig. 3(a). Suppose $X \in \mathbb{R}^{c_{in} \times h \times w}$ is the input feature in which $\emph{h}$ and $\emph{w}$ denote the spatial dimensions and $c_{in}$ is the number of channels. OctConv first decomposes $\emph{X}$ into two parts, one is HF $X^{H} \in \mathbb{R}^{(1 - \alpha_{in})c_{in} \times h \times w}$, and the other is LF $X^{L}{\in \mathbb{R}}^{\alpha_{in}c_{in} \times 0.5h \times 0.5w}$. The parameter $\alpha_{in}$ adjusts the number of channels in $X^{H}$ and $X^{L}$. Then, the LF and HF features are processed by convolution and the interaction between two frequencies will be carried out through pooling and up-sampling operations. The process of OctConv can be expressed by the following equations:

\begin{equation}
	\begin{aligned}	
Y^{H} = f\left( X^{H};W^{H\rightarrow H} \right) + up\left( f\left( X^{L};W^{L\rightarrow H} \right),2 \right)
	\end{aligned}
	\end{equation}
\begin{equation}
	\begin{aligned}	
Y^{L} = f\left( X^{L};W^{L\rightarrow L} \right) + f\left( pool\left( X^{H},2 \right);W^{H\rightarrow L} \right)
	\end{aligned}
	\end{equation}
where $\emph{f(X;W)}$ represents the convolution with kernel $\emph{W}$, and $\emph{W}$ is divided into  $W^{H}$ and $W^{L}$ to convolve with  $X^{H}$ and $X^{L}$ respectively. $W^{H}$ can be further divided into $W^{H\rightarrow H}$ and $W^{L\rightarrow H}$ for intra- and inter-frequency processing. Similarly, $W^{L}$ can also be divided into $W^{L\rightarrow L}$ and $W^{H\rightarrow L}$. This process can realize the communication of LF and HF information. To deal with the mismatch between spatial scales of $X^{H}$ and $X^{L}$, $\emph{pool(X,2)}$ and $\emph{up(X,2)}$ are used. $\emph{pool(X,2)}$ represents average pooling with kernel size 2 $\times$ 2 and stride 2, and  $\emph{up(X,2)}$ is an up-sampling operation by a factor of 2. Through the above operations, the output HF features $Y^{H}{\in \mathbb{R}}^{(1 - \alpha_{out})c_{out} \times h \times w}$ and LF features $Y^{L}{\in \mathbb{R}}^{\alpha_{out}c_{out} \times 0.5h \times 0.5w}$ can be obtained. The parameter $\alpha_{out}$ is a parameter adjusts the output channels $c_{out}$.

Based on OctConv, the proposed FSM is shown in Fig. 3(b). First, a 1$\times$1 OctConv ($\alpha_{in}$=0, $\alpha_{out}$=0.5) is utilized to divide the input feature into LF and HF parts. Then a 3$\times$3 OctConv ($\alpha_{in}$=0.5 and $\alpha_{out}$=0.5) is applied to obtain the LF and HF features (denoted by $Y^L$ and $Y^H$). Next, a 1$\times$1 OctConv ($\alpha_{in}$=0.5, $\alpha_{out}$=0) is used to fuse the LF and HF features into a whole for the subsequent down-sampling or up-sampling operation. At last, a residual connection is utilized to integrate the input feature with the output of the last OctConv, so that important information is not lost in FSM. 

\subsection{Cross-scale Feature Fusion Module}

\begin{figure}[h]
\centering
\includegraphics[width=0.9\columnwidth]{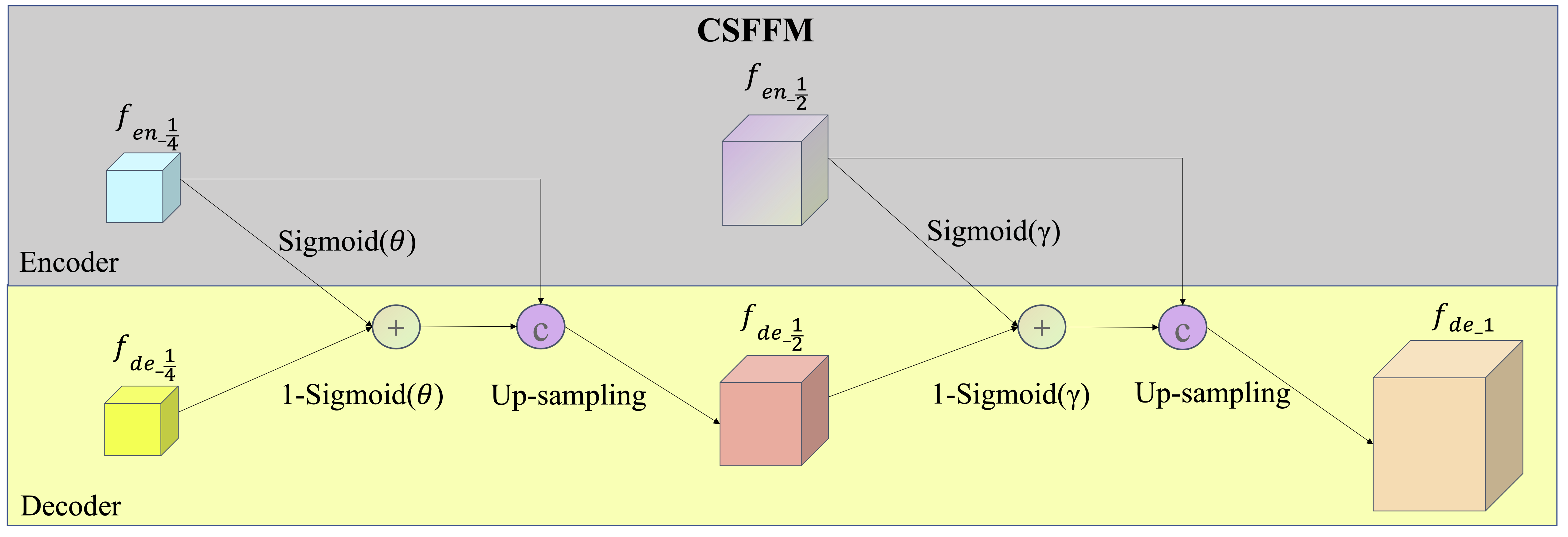} 
\caption{The architecture of the cross-scale feature fusion module (CSFFM).}
\label{fig1}
\end{figure}

From the analysis and observation in previous sections, we know that the image features with different scales exhibit different characteristics. The image feature with large scale contains fine structures such as clear edges and textures. However, with the down-sampling of feature scale, the fine structures will gradually degenerate and only the coarse structures (such as the rough contours) in the image are left. To make up for the information loss caused by down-sampling and achieve a cross-scale feature fusion, a CSFFM is proposed in our study. The CSFFM is based on the idea of adaptive mix-up operation~\cite{46} and its specific process is shown in Fig. 4. The process can be formulated as the following equations:

\begin{equation}
	\begin{aligned}	
f_{de\_\frac{1}{2}} = up\left( \left\lbrack {\begin{pmatrix}
{Sigmoid(\theta)*f_{en\_\frac{1}{4}}} \\
{+ \left( {1 - Sigmoid(\theta)} \right)*f_{de\_\frac{1}{4}}} \\
\end{pmatrix},f_{en\_\frac{1}{4}}} \right\rbrack \right)~
	\end{aligned}
	\end{equation}

\begin{equation}
	\begin{aligned}	
f_{de\_ 1} = up\left( \left\lbrack {\begin{pmatrix}
{Sigmoid(\gamma)*f_{en\_\frac{1}{2}} +} \\
{\left( 1 - Sigmoid(\gamma) \right)*f_{de\_\frac{1}{2}}} \\
\end{pmatrix},f_{en\_\frac{1}{2}}} \right\rbrack \right)
	\end{aligned}
	\end{equation}
where $f_{en\_\frac{1}{i}}$ and $f_{de\_\frac{1}{i}}$ represent the features of 1/$\emph{i}$ scale in encoder and decoder stages (the value of $\emph{i}$ is 1, 2 and 4 in our study), $\theta$ and $\gamma$ are the parameters optimizable by network, $\emph{up}$ represents the up-sampling operation and [] denotes the concatenation. CSFFM not only realizes the fusion of features at different scales but also connects the features of encoder and decoder, so that the important features of input image can be retained.

\subsection{Contrastive Learning Module}

The main idea of contrastive learning is to pull the positive paired samples together while push negative paired samples far apart in a feature space. In our study, a contrastive learning module (CLM) is proposed to regularize the HF features in decoder stage to get better restored images. According to Fig. 2, we can see that each CLM leverages three different features at the same scale to construct the positive and negative pairs for contrast. Here, we take the HF features obtained by the encoder stage, output sharp image and decoder stage as negative samples, positive samples and anchors, respectively. The reasons for this design are two-fold. First of all, the HF features in each scale of encoder stage are mainly captured from the blurry image, so the information contained in them is unclear and undesirable. Secondly, since the output sharp image of backbone is closest to ground-truth, its HF features can be considered as guidance for the intermedia features in the decoder stage. Through the CLMs at multiple scales and the loss function associated with them, the adverse information in the HF features of blurry image can be effectively suppressed.

\subsection{Loss Functions}

\subsubsection{Multi-scale Consistent Loss for Low-frequency Features}

Inspired by cycle-consistency, we minimize the $\mathcal{L}_{1}$ distance between the LF features in encoder stage and those generated by the output sharp image, so that the LF features can be maintained during deblurring. Therefore, the multi-scale loss for LF features can be expressed by:

\begin{equation}
	\begin{aligned}	
\mathcal{L}_{low} = min{\sum\limits_{k = 0}^{2}\left| \middle| f_{k}^{en\_ low} - f_{k}^{low} \middle| \right|_{1}}
	\end{aligned}
	\end{equation}
where $\emph{k}$ represents scale level, $f_{k}^{en\_ low}$ and $f_{k}^{low}$ represent the $\frac{1}{2^{k}}$ scale LF features got by encoder and output sharp image. Here, we should point out that since the input blurry image and output sharp image are both decomposed into multi-scale LF and HF components by the same network, the consistent loss of LF features in Eq. (5) will lead the diversity of blurry and sharp images to be mainly reflected by their HF components at each scale, which could facilitate our contrastive learning module for HF features.

\subsubsection{Multi-scale Contrastive Loss for High-frequency Features}

Considering all CLMs at different scales, the multi-scale contrastive loss ($\mathcal{L}_{high}$) for regularizing the HF features in our network can be expressed by the following equation:

\begin{equation}
	\begin{aligned}	
\mathcal{L}_{high} = {\mathit{\min}{\sum\limits_{k = 0}^{2}{\frac{\mathcal{L}_{1}\left( f_{k}^{anchor},f_{k}^{positive} \right)}{\mathcal{L}_{1}\left( {f_{k}^{anchor},f_{k}^{negative}~} \right)}~~~~}}}
	\end{aligned}
	\end{equation}
where $\emph k$ represents the scale level of feature, $f_{k}^{anchor}$, $f_{k}^{positive}$ and $f_{k}^{negative}$ represent the $\frac{1}{2^{k}}$ scale HF features obtained by decoder, output sharp image and encoder, respectively. $\mathcal{L}_{1}$ represents the $\mathcal{L}_{1}$-distance. Minimizing Eq. (6) can pull the $f_{k}^{anchor}$ and $f_{k}^{positive}$ together and push the $f_{k}^{anchor}$ and $f_{k}^{negative}$ apart.

\subsubsection{The Final Loss of MSFS-Net}

At last, the total loss function used to train our MSFS-Net can be defined as:

\begin{equation}
	\begin{aligned}	
\mathcal{L}_{total} = {\lambda_{1}\mathcal{L}}_{high} + {\lambda_{2}\mathcal{L}}_{low} + min\left| \middle| I~ - ~G \middle| \right|_{1}
	\end{aligned}
	\end{equation}
where $\lambda_{1}$=$\lambda_{2}$ are set as 0.05 by experiment, $\emph{I}$ represents the output of our network and $\emph{G}$ is the ground-truth, $\mathcal{L}_{1}$ norm is applied to minimize the loss between the recovered image and ground-truth.

\subsection{Comparison with Other Methods}

To highlight the novelty of the proposed model, we compare MSFS-Net with some related methods. First, though the frequency separation has been adopted in some works to deal with image restoration problem such as super-resolution~\cite{48,49}  and deraining~\cite{51}, they only decomposed different frequency information from image features at a specific size. Thus, the LF and HF image features at different scales are neglected in them. For the methods which considered the frequency information of images in deblurring task~\cite{9,13}, they either only focused on HF features or indiscriminatingly treated different frequency features of an image with the same strategy. Hence, they are still different from our proposed method. Besides, unlike some of the aforementioned methods that embed discrete cosine transform (DCT), discrete wavelet transform (DWT) and their inverse operations into the network for frequency analysis, the pure convolution based network architecture of our MSFS-Net can avoid information interchanges between the frequency and spatial domains, which makes the information propagate more smoothly. Second, contrastive learning has also been employed in some image-to-image regression tasks~\cite{38,39}. However, these methods merely leveraged the contrastive learning to regularize the final output rather than the intermedia layers of the network. The different frequencies of multi-scale image features are ignored in them. The last technique related to our work is the perceptual loss~\cite{52} which also utilizes a multi-layer network to extract the features of network’s output. Nevertheless, the differences between perceptual loss and our work are still apparent. The aim of perceptual loss is to measure the visual difference between the network's output and the ground-truth by features extracted from a pre-trained deep neural network (i.e., VGG~\cite{53}). Hence, it cannot be adopted to constrain the features obtained by intermedia layers of backbone network. Furthermore, perceptual loss also overlooks the different frequency information of the image.

\section{Experiments}

\subsection{Dataset and Implementation Details}

We use the training set in GoPro dataset to train our model and the test set to validate our model. Besides, HIDE~\cite{54} and RealBlur~\cite{55} datasets are also employed to evaluate our model. GoPro dataset contains 3214 pairs of blurry and sharp images, in which the training and test sets consist of 2103 and 1111 pairs, respectively. HIDE dataset consists of 8422 pairs of blurry and sharp images and these images are carefully selected from 31 high-fps videos. RealBlur dataset consists of two subsets: RealBlur-J and RealBlur-R. For implementation details, the AdamW optimizer with parameter setting as $\beta_1$=0.9, $\beta_2$=0.9, $\epsilon$=1e-8 is used to optimize our network. The epochs and batch size are set as 3000 and 4 respectively. The initial learning rate is set as 1e-4 and decreased by the factor of 0.5 at every 500 epochs. 

\subsection{Quantitative and Qualitative Evaluation}

\subsubsection{GoPro Dataset}

To demonstrate the effectiveness of the proposed MSFS-Net, we compare the performance of our method with some state-of-the-art algorithms on GoPro dataset. The quantitative comparison result is listed in Table 1, and some visual comparisons are shown in Fig. 5. In our experiment, the PSNR and SSIM results of all comparison methods are directly quoted from their corresponding literatures. Recently,~\cite{64} have shown that Test-time Local Converter (TLC) can effectively reduce the inconsistency of train-test information distributions and improve the performance of image restoration without any model fine-tuning. Thus, we also combine TLC with our proposed MSFS-Net (denoted as MSFS-Net-Local) to compare its deblurring performance with some improved versions of other methods.

\begin{table}[h]
\scriptsize
\centering
\caption{Performance comparison on GoPro and HIDE datasets.}
\label{Tab03}
\begin{tabular}{ccccc}
\toprule
\multirow{2}*{\textbf{Methods}} &\multicolumn{2}{c}{\textbf{GoPro}} &\multicolumn{2}{c}{\textbf{HIDE}}\\

 &\textbf{PSNR}  & \textbf{SSIM } & \textbf{PSNR}  & \textbf{SSIM }  \\
\midrule
DeblurGAN~\cite{19}   &28.70	&0.858 &24.51	& 0.871\\

MSCNN~\cite{7}  &29.08	&0.914 &25.73	&0.874\\

~\cite{17} &29.19	&0.931 &-	&-\\

DeblurGAN-v2~\cite{20} &29.55	&0.934 &26.61	&0.875\\

~\cite{58} &29.81	&0.937 &-	&-\\

DMPHN~\cite{16} &30.21	&0.935 &29.09	&0.924\\

SRN~\cite{10} &30.26	&0.934 &28.36	&0.915\\

~\cite{13} &30.31	 &0.920 &-	&-\\

~\cite{23} &30.92 &0.942 &29.11	&0.913\\

DBGAN~\cite{60} &31.10	&0.942 &28.94	&0.915\\
 
MT-RNN~\cite{59} &31.15	&0.945 &29.15	&0.918\\

SDWNet~\cite{9} &31.26	&0.966 &28.99	&0.957\\

~\cite{63} &31.66	&0.948 &29.77	&0.922\\

MIMO-UNet~\cite{8} &31.73	 &0.951  &-	&-\\

RADN~\cite{24} &31.76	&0.953 &-	&-\\

~\cite{56}  &31.79	&0.949  &-	&-\\

~\cite{61} &31.85	&0.948 &29.98	&0.930\\

SPAIR~\cite{57} &32.06	&0.953  &30.29	&0.931\\

~\cite{25} &32.50	& 0.958 &30.55	&0.935\\

MPRNet~\cite{62} &32.66	&0.959 &30.96	&0.939 \\

HINet~\cite{15} &32.71	&0.959 &30.33 &0.932\\

\textbf{MSFS-Net}  &\textbf{32.73}	&\textbf{0.959} &\textbf{31.05}	&\textbf{0.941}\\

\cline{1-5}
MIMO-UNet++~\cite{8} &32.68	 &0.959  &-	&-\\

HINet-Local~\cite{64}	&33.08	&0.962  &30.66	&0.936\\

Whang et al.-SA~\cite{63}	&33.23	&0.963 &30.07	&0.928\\

MPRNet-Local~\cite{64}	&33.31	&0.964 &31.19	&0.942\\

\textbf{MSFS-Net-Local}  &\textbf{33.46}	&\textbf{0.964} &\textbf{31.30}	&\textbf{0.943}\\

\bottomrule
\end{tabular}
\end{table}

From Table 1, we can see that the deblurring performance of our method is superior to other state-of-the-art methods. The advantage of MSFS-Net can be attributed to the following reasons. First, different from some comparison methods that only adopt a simple skip connection mechanism to concatenate the features with the same scale in encoder and decoder~\cite{8,10}, the CSFFM in our model can better fuse the features of different scales and stages (i.e., encoder and decoder). Second, the methods in~\cite{7,19} integrate the multi-scale image features by some sophisticated network structure and modules. However, they treat the image as a whole and neglect the characteristics of different image frequencies. Thus, their performance is inferior to our model which makes full use of the LF and HF information separated by FSM. At last, the method in~\cite{13} considers the frequency information in image deblurring problem. But it only focuses on the HF image features. Although SDWNet~\cite{9} utilizes the DWT for image frequency separation, it indistinguishably processes the LF and HF image features using the same network. Hence, the deblurring results of above two methods are worse than our MSFS-Net which utilizes different strategies (i.e. contrastive learning and consistent loss) to handle muti-scale HF and LF features separately. From Table 1, we can also see that TLC promotes the performance of our MSFS-Net and MSFS-Net-Local outperforms the improved versions of some other methods. Through the visual comparison in Fig. 5, the  superiority of our MSFS-Net over other methods can be intuitively demonstrated.

In Fig. 6, we show the entropy distributions of LF and HF components obtained by sharp images and deblurred images of our MSFS-Net. Through comparing the results with those in Fig. 1, we can see that our proposed method can effectively narrow the gap between the sharp and blurry images. That is, the Jenson-Shannon divergencies between entropy distributions of different frequencies at each scale are much smaller than those in Fig. 1.

\begin{figure}[h]
\centering
\includegraphics[width=0.9\columnwidth]{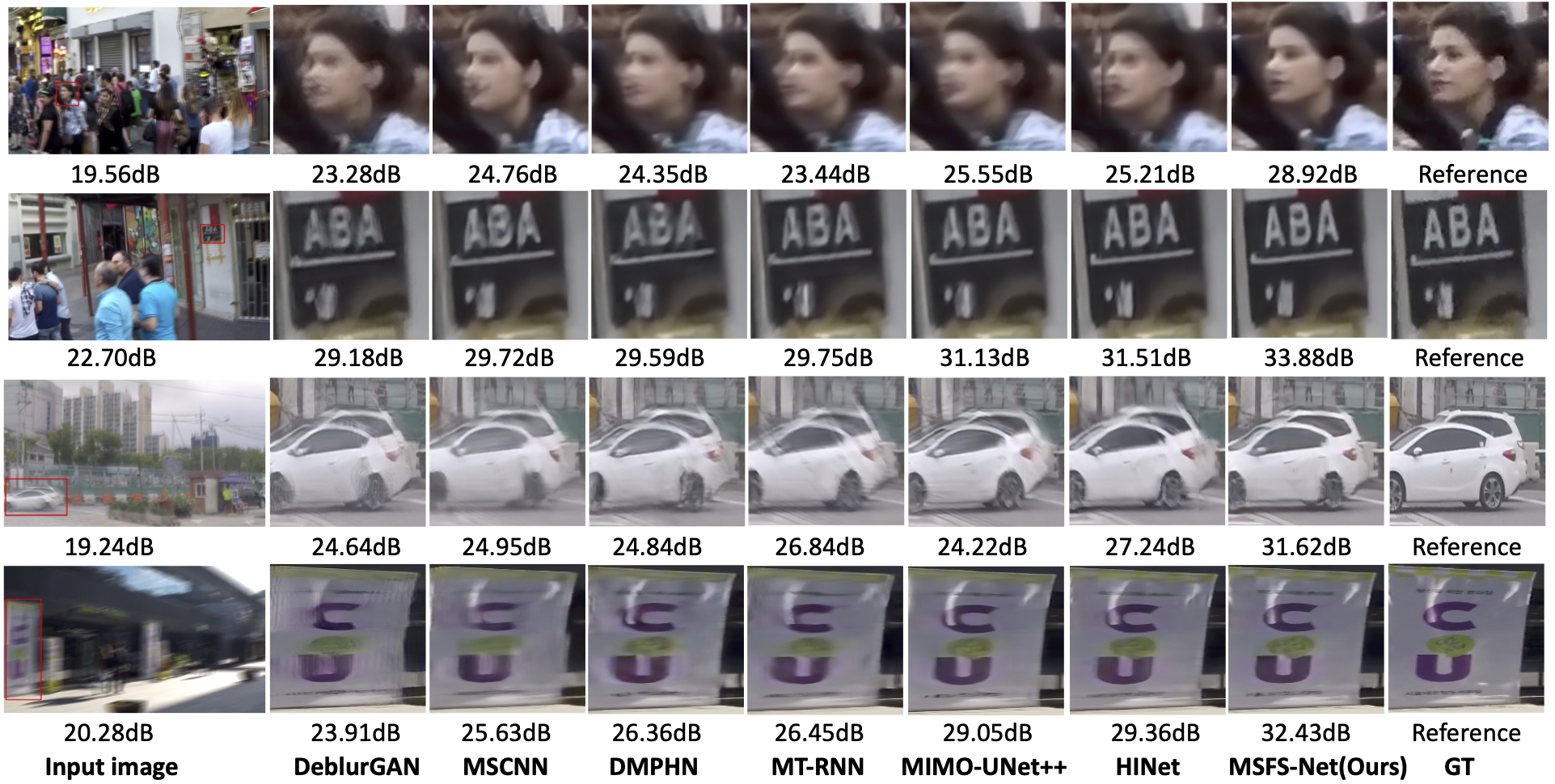} 
\caption{Visual comparison of the deblurring results on GoPro dataset.}
\label{fig1}
\end{figure}

\begin{figure}[h]
\centering
\includegraphics[width=0.8\columnwidth]{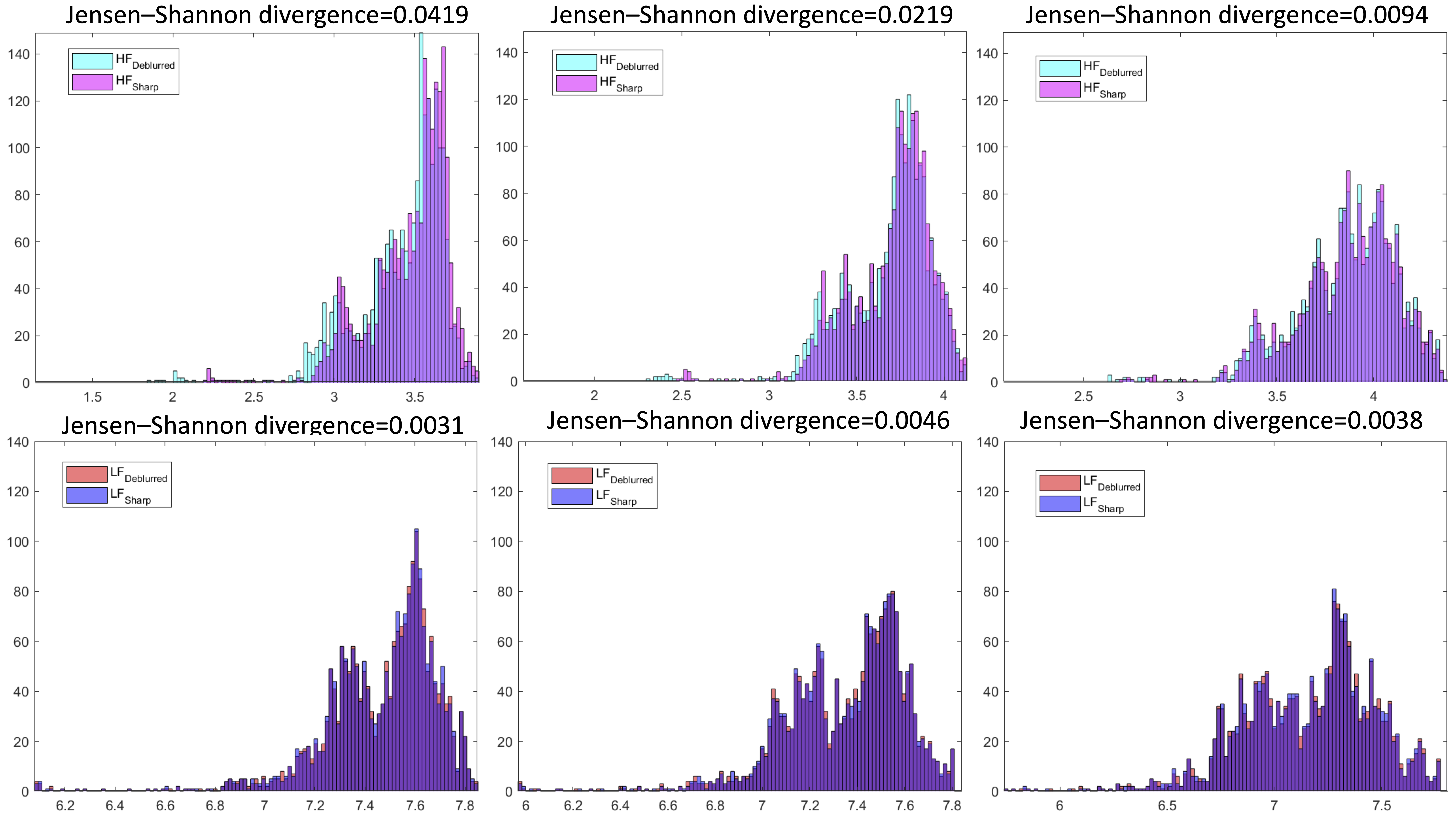} 
\caption{The distributions of entropy obtained by samples in GoPro dataset. Top: From left to right are the distributions of entropy obtained by HF components of sharp ($HF_{Sharp}$) and deblurred ($HF_{Deblurred}$) images at original, 1/2 and 1/4 scales. Down: From left to right are the distributions of entropy obtained by LF components of sharp ($LF_{Sharp}$) and deblurred ($LF_{Deblurred}$) images at original, 1/2 and 1/4 scales.}
\label{fig1}
\end{figure}

\subsubsection{HIDE and RealBlur Datasets}

Following some other works, we also evaluate our GoPro-trained MSFS-Net on HIDE and RealBlur datasets to test its generalization ability. From the quantitative and visual comparison results in Table 1 and Fig. 7, it can be seen that the proposed method achieves the best deblurring result on HIDE dataset. Similarly, the advantage of our MSFS-Net for handling real blurry images is justified in Table 2. Here, we should note that TLC cannot improve the performance of our GoPro trained model when it is directly applied on RealBlur dataset. This may due to the blurry images in RealBlur are captured in real scenario rather than synthesized from video. Thus, their characteristics are different from the training samples in GoPro. However, once the training of our model is conducted on RealBlur, MSFS-Net-Local outperforms the original MSFS-Net and some other methods.

\begin{figure}[h]
\centering
\includegraphics[width=0.88\columnwidth]{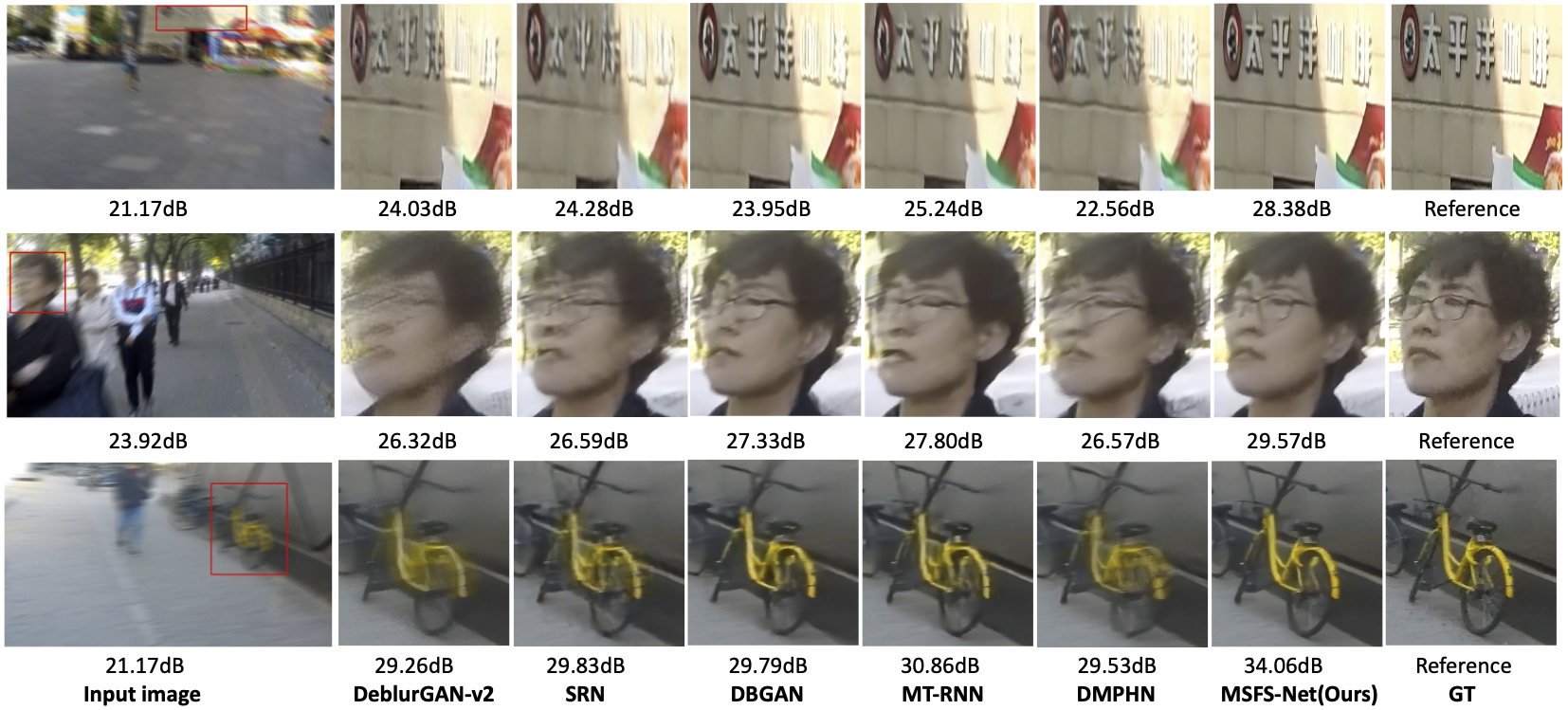} 
\caption{Visual comparison of the deblurring results on HIDE dataset.}
\label{fig1}
\end{figure}

\begin{table}[h]
\scriptsize
\centering
\caption{Performance comparison on RealBlur dataset under two different settings: 1). applying our GoPro trained model directly on the RealBlur set. 2). Training and testing on RealBlur data where methods are denoted with symbol *.}
\label{Tab03}
\begin{tabular}{ccccc}
\toprule
\multirow{2}*{\textbf{Methods}} &\multicolumn{2}{c}{\textbf{RealBlur-R}} &\multicolumn{2}{c}{\textbf{RealBlur-J}}\\

 &\textbf{PSNR}  & \textbf{SSIM } & \textbf{PSNR}  & \textbf{SSIM }  \\
\midrule
MSCNN~\cite{7}	&32.51	&0.841	&27.87	&0.827\\

DeblurGAN~\cite{19}	&33.79	&0.903	&27.97	&0.834\\

DeblurGAN-v2\cite{20} &35.26	&0.944	&28.70	&0.866 \\

~\cite{17}	&35.48	&0.947	&27.80	&0.847\\

SRN~\cite{10}	&35.66	&0.947	&28.56	&0.867 \\

DMPHN~\cite{16}	&35.70	&0.948	&28.42	&0.860\\

MIMO-UNet~\cite{8}	&35.47	&0.946	&27.76	&0.863 \\

SDWNet~\cite{9} &35.85	&0.948 &28.61	&0.867\\

\textbf{MSFS-Net}  &\textbf{36.02}	&\textbf{0.959} &\textbf{28.97} &\textbf{0.908}\\

\textbf{MSFS-Net-Local}  &\textbf{36.01}	&\textbf{0.958} &\textbf{28.89} &\textbf{0.906}\\

\cline{1-5}

DeblurGAN-v2*~\cite{20} &36.44	&0.935	&29.69	&0.870\\

SDWNet*~\cite{9} &38.21	&0.963 &30.73	&0.896\\

\textbf{MSFS-Net*}  &\textbf{38.26}	&\textbf{0.972} &\textbf{30.89} &\textbf{0.929}\\

\textbf{MSFS-Net-Local*}  &\textbf{38.87}	&\textbf{0.974} &\textbf{31.53} &\textbf{0.932}\\

\bottomrule
\end{tabular}
\end{table}

\subsection{Ablation Study and Analysis}

In this section, we conduct several experiments to evaluate the effectiveness of the components proposed in our MSFS-Net. Through discarding each component in our model, we can get five new network structures (MSFS-Net w/o FSM but w/ CLM or Consistency, w/o CSFFM, w/o CLM and w/o Consistency). In order to fairly compare their performance, we use the same parameter settings to train these networks. The results of ablation experiment on GoPro dataset are shown in Table 3. First, neither CLM nor Consistency constraint can achieve satisfied deblurring result when we don’t decompose the image features into different frequency by FSM. On the one hand, the energy of an image is mostly concentrated in its LF components. Thus, if we neglect different frequency information and treat the image features as a whole, the contrastive loss in Eq. (6) will be dominated by the very similar LF components of blurry and sharp images. As a result, the difference between HF components of blurry and sharp images, which is very crucial for image deblurring, will be overlooked. On the other hand, imposing the Consistency constraint on the unseparated frequency features will also prevent our model from restoring the HF details during the deblurring. Second, the removal of CSFFM deteriorates the performance of our model. This is due to CSFFM can make connections between features at different stages and scales, which results in a better information fusion. Finally, we can see that the absence of Consistency constraint or CLM for separated frequency features also has an adverse impact on the deblurring result of the proposed network. This justifies that constraining the LF and HF features in the intermediate layers with consistency criterion and contrastive learning are both important for improving the deblurring performance.

\begin{table}[h]
\scriptsize
\centering
\caption{Comparison of different ablations of MSFS-Net on GoPro dataset.}
\label{Tab03}
\begin{tabular}{ccccccc}
\toprule
\textbf{FSM} &\textbf{CSFFM}   &\textbf{CLM}   &\textbf{Consistency} &PSNR &SSIM \\
\midrule
$\times$  & $\checkmark$  & $\checkmark$ &$\times$   &30.01	&0.908\\
$\times$  & $\checkmark$  & $\times$ & $\checkmark$  &30.13	&0.910\\
$\checkmark$  & $\times$   &$\checkmark$ &$\checkmark$  &30.22	&0.891 \\
$\checkmark$ &$\checkmark$ &$\times$  & $\checkmark$ &31.08	&0.911 \\
$\checkmark$  & $\checkmark$    &$\checkmark$ & $\times$  &31.54	&0.913 \\
$\checkmark$  & $\checkmark$   & $\checkmark$ & $\checkmark$ &\textbf{32.73}	&\textbf{0.959}\\

\bottomrule
\end{tabular}
\end{table}

\section{Conclusion}

In this work, we propose a multi-scale frequency separation network (MSFS-Net) for image deblurring. In order to make the network take full advantage of the LF and HF features, we propose FSM to separate  features into different frequency. At the same time, to make the features of different scales communicate with each other without losing information, CSFFM is proposed to realize feature connection. Finally, the cycle-consistency and contrastive learning strategies are designed by analyzing the different characteristics of LF and HF features between blurry and sharp images.  Experiments on three datasets show that MSFS-Net achieves good results in image deblurring task.

In the future, we will incorporate some novel backbone (such as Transformer) into our network to improve its feature learning power and apply our model to other image restoration tasks (such as deraining, dehazing and denoising) to test its generalization ability.

\bibliography{our}
\nobibliography{aaai22}

\end{document}